\documentclass[10pt,twocolumn,letterpaper]{article} 

\usepackage{avss}
\usepackage{times}
\usepackage{epsfig}
\usepackage{graphicx}
\usepackage{amsmath}
\usepackage{amssymb}

\usepackage{gensymb}
\usepackage{float}
\usepackage{placeins}
\usepackage{pgf}
\usepackage{multirow}
\usepackage{bigdelim}
\usepackage{arydshln}
\usepackage{stmaryrd}
\usepackage{mathtools}
\usepackage{tikz}

\usepackage{pifont}

\usepackage{bibspacing}

\newcommand{\tablebottom}{\noalign{\hrule height 0.35ex}}
\newcommand{\tabletop}{\noalign{\hrule height 0.35ex}} %
\newcommand{\tablecbottom}[1]{\cline{#1}}
\newcommand{\tablectop}[1]{\cline{#1}}

\renewcommand{\checkmark}{\ding{51}}%
\newcommand{\crossmark}{\ding{55}}%

\def\x{{\mathbf x}}
\def\normset{\mathcal{X}_{\text{norm}}}
\def\task{\mathcal{T}}
\def\softmax{{\operatorname{softmax}}}

\newcommand{\groupcell}[4]{\multirow{#1}{*}{}\ldelim\{{#1}{#2}[\parbox{#3}{#4}\ ]}

\newcommand{\copyrightnotice}[1]{%
	\begin{tikzpicture}[remember picture,overlay]
		\node[anchor=south,yshift=40pt] at (current page.south) {\parbox{\dimexpr\textwidth-\fboxsep-\fboxrule\relax}{\footnotesize #1}};
	\end{tikzpicture}\vspace{-12pt}%
}

\usepackage[pagebackref=true,breaklinks=true,letterpaper=true,colorlinks,bookmarks=false]{hyperref}

\avssfinalcopy % *** Uncomment this line for the final submission

\def\avssPaperID{21} % *** Enter the AVSS Paper ID here

\ifavssfinal\fi

\begin{document}

%%%%%%%%% TITLE
\title{Fine-grained anomaly detection via multi-task self-supervision}

\author{Loïc Jézéquel $^{1, 2}$\hspace*{1cm} Ngoc-Son Vu $^1$\hspace*{1cm} Jean Beaudet $^2$\hspace*{1cm}  Aymeric Histace  $^1$ \\ \\
$^1$ ETIS UMR 8051 (CY Cergy Paris Université, ENSEA, CNRS) F-95000\\ $^2$ Idemia Identity \& Security, 95520 Osny France \\
{\tt\small \{loic.jezequel, son.vu, aymeric.histace\}@ensea.fr}
}
\maketitle
\copyrightnotice{978-1-6654-3396-9/21/\$31.00 ©2021 IEEE}%
\thispagestyle{empty}

%%%%%%%%% ABSTRACT
\begin{abstract}
Detecting anomalies using deep learning has become a major challenge over the last years, and is becoming increasingly promising in several fields. The introduction of self-supervised learning has greatly helped many methods including anomaly detection where simple geometric transformation recognition tasks are used. However these methods do not perform well on fine-grained problems since they lack finer features. By combining both high-scale shape features and low-scale fine features in a multi-task framework, our method greatly improves fine-grained anomaly detection. It outperforms state-of-the-art with up to 31\% relative error reduction measured with AUROC on various anomaly detection problems including one-vs-all, out-of-distribution detection and face presentation attack detection.
\end{abstract}

%%%%%%%%% BODY TEXT
\label{sec:intro}

Detecting anomalies straying apart from a well-defined normal situation has always been a major challenge in many fields such as video surveillance \cite{Sultani2018,Zhu2020}, intrusion detection \cite{netintrusion2019}, fraud detection \cite{onlinefraud2018}, medical imaging \cite{LUNDERVOLD2019102} and more recently adversarial attack detection \cite{advtrainexdetection2018}. Deep visual anomaly detection has been introduced to tackle this problem and has proven to be more robust and reliable than classical binary classification. Rather than directly try to discriminate anomalies from normal samples, we only learn the normal class boundary and deem as anomalous any observation outside.

Recently, the introduction of self-supervised learning has greatly improved many one-class anomaly detection learning methods. It enables to discriminate anomalies from normal samples by learning to solve simple tasks such as geometric transformation classification \cite{Golan2018DeepAD}. However, even if this approach has greatly improved anomaly detection performance, it still suffers from limitations on more challenging problems with local and fine-grained differences between anomalies and normal samples.

In this given context, our main contributions in this paper are the following:
\begin{itemize}
	\item We improve the detection of fine-grained anomalies by independently solving in a multi-task self-supervised fashion high-scale geometric task and low-scale jigsaw puzzle task.
	\item We validate the efficiency of the proposed method using an exhaustive protocol for anomaly detection on one-vs-all, out-of-distribution detection and anti-spoofing problems.
	\item The proposed method obtains better overall results with up to 31\% AUROC relative improvement from state of the art methods.
\end{itemize}

\section{Related work}

\subsection{Anomaly detection}

The main goal in anomaly detection is to classify a sample as normal or anomalous. Formally, we predict $P(\x\in\normset)$ for an observation $\x$ and a normal (or positive) class $\normset$. In practice, a proxy anomaly score function $s_a(\x)$ is usually estimated instead. The anomalous (or negative) class is then defined implicitly as the complementary of the normal class in image space.
We can generally categorize anomalies into three families:

\begin{enumerate}
	\item \textbf{Object anomaly}: any object which is not included in the positive class, e.g., a cat is an object anomaly in regards to dogs.
	\item \textbf{Style anomaly}: observations representing the same object as the positive class but with a different style or support, e.g., a realistic mask or a printed face represent faces but with a visible different style.
	\item \textbf{Local anomaly}: observations representing and sharing the same style as the positive class, however a localized part of the image is different. Most of the time, these anomalies are the superposition of two generative processes, e.g., a fake nose on a real face is a local anomaly.
\end{enumerate}

Usually, we assume in anomaly detection that only normal samples are available during training, meaning that most methods are part of one-class learning scheme. The first introduced methods simply used a pre-trained neural network to extract features, on which a classical algorithm such as One-Class SVM \cite{Schlkopf1999SupportVM} (OCSVM) or Isolation Forest \cite{isolationforest2009} (IF) were trained.

There have been also semi-supervised anomaly detection methods such as DeepSAD \cite{deepsad2020} or deviation networks \cite{Pang2019DeepAD} where we assume some of the anomalies representing a few modes are available. These methods can achieve better accuracy on borderline cases given enough diverse anomalies, which is often less manageable in practice. In particular, these two methods directly learn representations by minimizing the distance of normal sample features to an hypersphere center, while maximizing the distance to the anomalies. It follows the compactness principle, where we minimize the normal class representations variance and maximize the inter-class representations variance.

\subsection{Self-supervised learning}

Self Supervised Learning (SSL) is a part of representation learning, where we want to learn useful and general representations from an unlabeled dataset $\mathcal{X}=\{\x_i\}_1^N$. We can then use the learned features for a different task such as classification.

We learn representations by solving from the data an auxiliary task $\task$, which is often unrelated to the final one. Therefore SSL consists of two steps:
\begin{enumerate}
	\item Generating a labeled dataset $\mathcal{X_\task}$ aligned with $\task$, which for classification is usually done by applying $c$ transformations $T_j$ to our unlabeled samples 
	\begin{equation}
		\mathcal{X}_\task=\{(T_j(\x_i), j)\}_{i,j}
	\end{equation}
	\item Training a classification or regression network on this generated labeled set.
\end{enumerate}
One of the final layers $\phi_\task$ can thus be used as a feature extractor. Some commonly used tasks are: 90° rotation prediction \cite{Gidaris2018UnsupervisedRL}, jigsaw puzzle \cite{jigsaw2016}, distortions \cite{examplar2014}, colorization \cite{Zhang2016ColorfulIC}, image inpainting \cite{Pathak2016ContextEF} or relative patches prediction \cite{patchpos2015}.

\subsection{SSL anomaly detection}
\label{subsec:sslad}

Very recently, SSL has been adapted to the one-class anomaly detection framework. First we learn to solve an auxiliary task $\task$ in a SSL fashion to obtain a pre-trained network $\phi_\task$. Then, to classify at inference time an observation $\x$ as anomalous or normal, we evaluate how well the network can solve the task. Indeed, the main assumption is that the network will perform relatively well on normal samples but will fail on anomalies. A task-independent metric $L$ is computed on the generated labeled samples to compute the anomaly score function:
\begin{equation}
	s_a(\x)=\left\{L(\phi_\task(T_i(\x)), i)|i\in\llbracket 1,c\rrbracket\right\}
\end{equation}

Unlike SSL, we are not directly interested in the intermediate features, but rather the final task outputs.

In \textbf{GeoTrans} \cite{Golan2018DeepAD}, the auxiliary task is to classify which geometrical transformation has been applied to the input. A set of 72 transformations including identity is randomly sampled over all possible compositions of translations, rotations and symmetries. At the end of training, 72 Dirichlet distributions respectively parameterized by $\tilde{\boldsymbol{\alpha}}_{i}$ are fitted over the normal class softmax outputs $\mathbf{y}\left(T_{i}(\x)\right)$ for each transformation. The log-likelihood can then be used during inference as the task-independent metric $L$:

\begin{equation}
	s_a(\x)=\sum_{i=1}^{72}\left(\tilde{\boldsymbol{\alpha}}_{i}-1\right) \cdot \log \mathbf{y}\left(T_{i}(\x)\right)
\end{equation}

In \textbf{MHRot} \cite{Hendrycks2019UsingSL}, the task is to simultaneously classify three types of transformations, each modeled by a softmax head: vertical translation, horizontal translations and 90° rotations. Accordingly, we are trying to predict the three following variables: vertical translations (0,$-t_y$,$+t_y$), horizontal translations (0,$-t_x$,$+t_x$) and 90° rotations (0°,90°,180°,270°).

During inference, we sum the three softmax of the known transformations for each transformation combination:

\begin{equation}
	\label{eq:mhrot_as}
	% \resizebox{1 \columnwidth}{!}
	% {
	% 	$\displaystyle s_a(x)=\sum\limits_{\mathclap{\substack{ r\in\{0, 90, 180, 270\} \\ s\in\{0, -t_x, +t_x\} \\  t\in\{0, -t_y, +t_y\}}}} \;\mathbf{y}_{rot}(T_{\scriptscriptstyle r,v,t}(x))_r+\mathbf{y}_{v}(T_{\scriptscriptstyle r,v,t}(x))_s+\mathbf{y}_{h}(T_{\scriptscriptstyle r,v,t}(x))_t$
	% }
	s_a(x)=\quad\sum\ \sum_{{\tiny \mathclap{\substack{ r\in\{0, 90, 180, 270\} \\ s\in\{0, -t_x, +t_x\} \\  t\in\{0, -t_y, +t_y\}}}}}\ \sum\quad \mathbf{y}(T_{\scriptscriptstyle r,s,t}(x))_{r,s,t}
\end{equation}

\section{Method overview}
\label{sec:method}
\subsection{Anomaly detection pretext task}

We present here a general rule of thumb regarding the choice of tasks for SSL anomaly detection. It is generally more restrictive than for simple representation learning \cite{asano2020critical}. 

\begin{figure}
	\begin{center}
	\resizebox{1.045\linewidth}{!}{
		\hspace{-14pt}
		\includegraphics{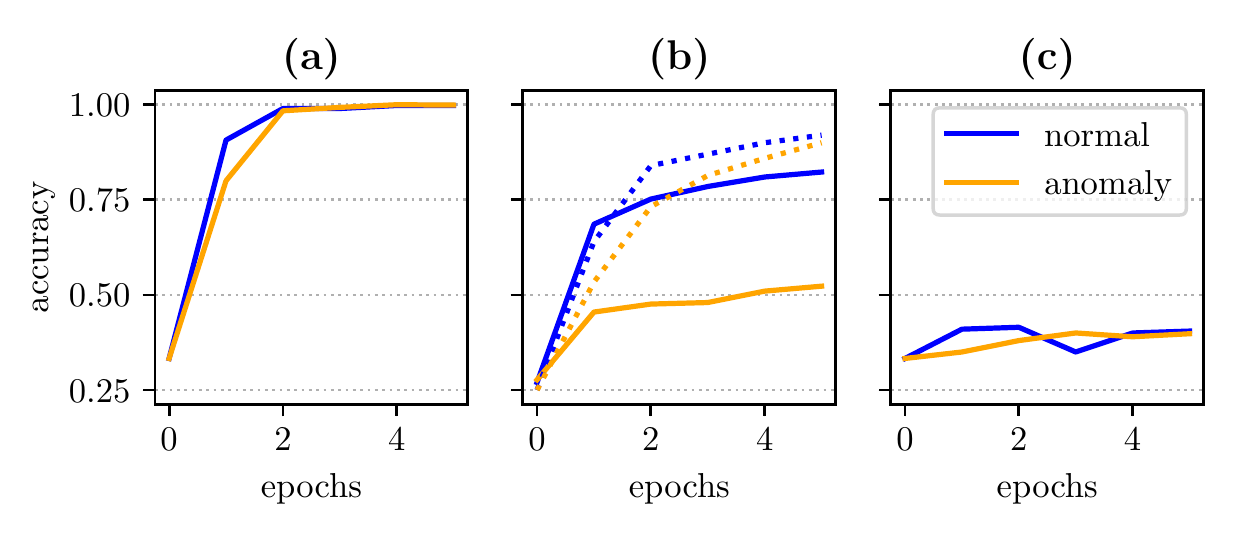}
	}
	\end{center}
	\caption{Tasks accuracy during training on coarse object AD (CIFAR-10 in plain line) and fine-grained AD (CaltechBirds in dotted line): \textbf{(a)} vertical translations, \textbf{(b)} 90° rotation, \textbf{(c)} unsolvable task}
	\label{fig:taskplot}
\end{figure}

Let $\task$ be a task  along its training loss $L_\task$. On the one hand, if the task is too hard on normal samples, meaning that the accuracy of our network remains close to random predictor throughout training (or that $\|\nabla_\phi L_\task \|$ is always small and that the minimum of $L$ is high), then no meaningful representation will be reached at convergence. This will also result in poor accuracy on anomalies (Fig.\ref{fig:taskplot}.c) and yield unpredictable results during anomaly detection.
On the other hand, if the task is too easy on normal samples, meaning that our model will converge to a perfect predictor in the first epochs, then the task loss will be minimized by many representations including trivial ones. Thus the network is more likely to learn such representations which will be unspecific to the normal class and encode very generic visual features. Since many anomalies will share these features, the task accuracy will be high on anomalies as well (Fig.\ref{fig:taskplot}.a).

To observe these effects, we train a network on several isolated tasks as described in Section \ref{subsec:sslad}. By monitoring its task classification accuracy on evaluation normal data and anomalous data during the first epochs, we empirically measure how well-suited a pretext task is for anomaly detection on a given dataset. We show that even though 90° rotation is more adapted than translations on coarse anomaly detection, it ultimately fails on fine-grained anomaly detection (Fig.\ref{fig:taskplot}.b). This confirms that basic geometric transformation recognition tasks, such as 90° rotations or translations, are only suited to simple object anomaly detection. Indeed, since these tasks are solvable accurately by learning high scale and shape features, it is unlikely the network will use finer characteristics that allow discriminating normal samples from more subtle anomalies.

\subsection{Method overview}

Finding a single task satisfying all the previous conditions is difficult, and must be highly dataset dependent. Therefore we resort to ensemble methods \cite{10.1007/3-540-45014-9_1} by allowing the network to learn $N$ tasks and merge their decision at inference. We learn richer features via multi-task learning \cite{Caruana1997}, by sharing a common representation across all tasks. Our model is accordingly composed of a main feature extractor network $\phi$ and $N$ dense layers $f_{\task_1},\cdots, f_{\task_N}$, where $f_{\task_i}(\phi(\x))$ is the output for the i\textsuperscript{th} task. 

\begin{figure*}[tb]
	\begin{center}
	%\vspace{-20px}
	\includegraphics[width=0.75\linewidth]{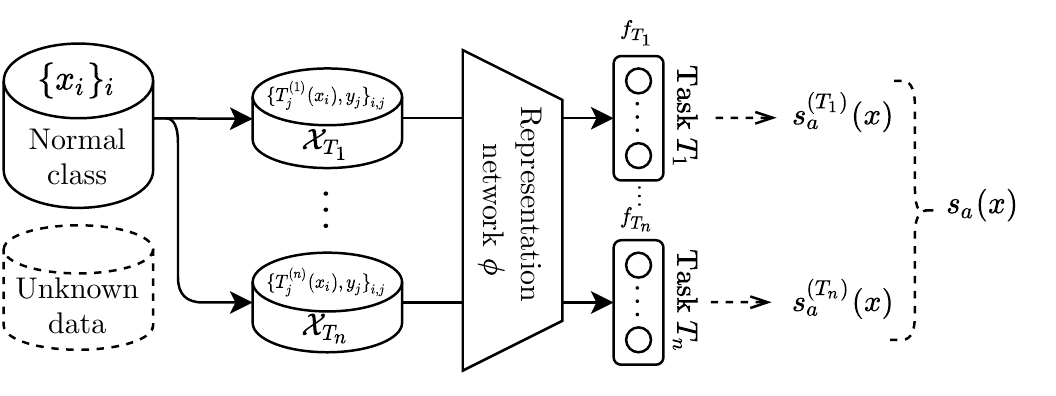}
	\end{center}
	\caption{Multi-task self-supervised anomaly detection. In dotted line are additional steps during inference}
	\label{fig:mtask}
\end{figure*}

During inference, we aggregate the anomaly scores of all tasks into the final anomaly detection score. The whole training and inference scheme is summarized in Figure \ref{fig:mtask}. For each classification task $\task_i$, the task-independent metric $L_{\task_i}$ chosen is the softmax score corresponding to the true known class and we sum up these scores using the mean:
\begin{equation}
	s_a(\x)=\frac{1}{N}\sum_{i=1}^N s_a^{(\task_i)}(\x)
\end{equation}

where $s_a^{(\task_i)}$ is the anomaly score of the i\textsuperscript{th} task:

\begin{equation}
	s_a^{(\task_i)}(\x)=\sum_j \softmax(\phi\circ f_{\task_i}(T_j^{(i)}(\x)))_j
\end{equation}

We note that there is a caveat using the mean as anomaly score: adding new tasks can have a negative impact on the model performance. In practice if the task is not well suited to the normal class, it will add significant noise to the anomaly score and ultimately harm the anomaly detection accuracy.

% Add here note difference with MHRot: each task is evaluated independently. => more efficient

To prevent our multi-task from being too easy on fine-grained problems, we introduce more challenging tasks. We choose here a simplified version of the jigsaw puzzle task. The jigsaw puzzle task consists in splitting an image into a grid of $n_h\times n_w$ patches, then randomly shuffling the different patches. The task is then to predict the original order of each patch. This task has proven in representation learning to provide a great challenge for extracting more local and finer features \cite{jigsaw2016}. To avoid trivial solutions and force our model to understand pieces neighborhood, we are careful to add a margin between each patch with a random small offset.

\begin{figure}[htb]
	\begin{center}
	\begin{minipage}[b]{0.32\linewidth}
		\centering
		\centerline{\includegraphics[width=\linewidth]{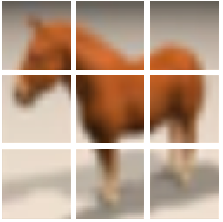}}
		\vspace{.2cm}
		\centerline{Permutation 1}\medskip
	\end{minipage}
	\begin{minipage}[b]{0.32\linewidth}
		\centering
		\centerline{\includegraphics[width=\linewidth]{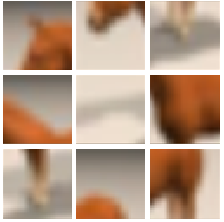}}
		\vspace{.2cm}
		\centerline{Permutation 2}\medskip
	\end{minipage}
	\begin{minipage}[b]{0.32\linewidth}
		\centerline{\includegraphics[width=\linewidth]{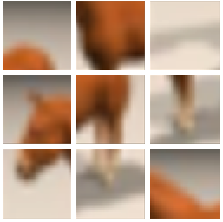}}
		\vspace{.2cm}
		\centerline{Permutation 3}\medskip
	\end{minipage}
	\end{center}
	\caption{Example of simplified jigsaw puzzle task for $k=3$}
	\label{fig:res}
\end{figure}

Since we chose the softmax truth as the task-independent metric, we need to re-frame it into a classification problem by considering each permutation as a single class. This would greatly increase our model complexity, effectively adding $(n_w\cdot n_h)!$ classes. Therefore, we only consider $k<(n_w\cdot n_h)!$ randomly chosen permutations including the identity permutation. This quantity $k$ becomes an additional parameter controlling the task difficulty.

The complete training loss for a single sample $\x$ becomes 

\begin{align}
	\textstyle
	L(\x)=
	\textstyle \sum_{i=1}^3 L_{CE}(\phi\circ f_\text{v}(T_i^{(\text{v})}(\x)), i)\,+ \nonumber\\
	\textstyle \sum_{i=1}^3 L_{CE}(\phi\circ f_\text{h}(T_i^{(\text{h})}(\x)), i)\,+ \nonumber\\
	\textstyle \sum_{i=1}^4 L_{CE}(\phi\circ f_\text{rot}(T_i^{(\text{rot})}(\x)), i)\,+ \nonumber\\
	\textstyle \sum_{i=1}^k L_{CE}(\phi\circ f_\text{puzz}(T_i^{(\text{puzz})}(\x)), i)
\end{align}

where $L_{CE}$ is the cross-entropy and $f_\text{v}, f_\text{h}, f_\text{rot}, f_\text{puzz}$ are respectively the dense layers for the vertical translations, horizontal translations, rotations and puzzle tasks. Compared to the MHRot anomaly score in Equation \ref{eq:mhrot_as} we evaluate each task \textit{independently}, which allows us to greatly reduce the required amount of network forward pass to compute the anomaly score during inference. As a consequence, our method inference step is roughly 10 times faster.

By combining the base geometrical transformation recognition task with the jigsaw task, we allow the model to learn high-scale shape features more suited toward object anomaly detection as well as low-scale fine features more suited toward style anomaly and local anomaly detection.

\section{Implementation details}

The geometrical transformation task is composed as in \cite{Hendrycks2019UsingSL} of horizontal translations, vertical translations and 90° rotations. As for the jigsaw puzzle task, we found best results with $n_w=n_h=3$ and $k=3$.

Regarding network architecture, we use a 16-4 WideResNet \cite{Zagoruyko2016WideRN} ($\approx10M$ parameters with a depth of 16) for the feature extractor network $\phi$, along with two dense softmax layers respectively of size $10$ for the geometrical transformation task and size $3$ for the jigsaw puzzle task. Each of these dense layers have a dropout rate of 0.3 \cite{Srivastava2014DropoutAS}. Training is performed under SGD optimizer with nesterov momentum \cite{Sutskever2013OnTI}, using a batch size of 32.

\section{Results}

\subsection{Evaluation protocol}

Until now, most of the anomaly detection literature have adopted the one-vs-all protocol to evaluate their method. In the one-vs-all protocol, we consider one class of a multi-class dataset, originally created for object recognition, as the normal class. All the other classes are then considered as anomalous, and we can in a leave-one-out cross-validation fashion evaluate the model on each possible normal class. The final reported result is the mean of each run.

Even though such datasets are easier to acquire and result in a highly multi-modal anomaly class, these might not be enough to fully evaluate anomaly detection methods. Indeed, these only cover coarse object anomalies which are now becoming too easy for state-of-the-art methods, and do not reflect realistic anomaly detection challenges.

\setlength{\dashlinedash}{.8pt}
\setlength{\dashlinegap}{1.8pt}

\begin{table}[tbh]
	\begin{center}
		\hspace{-25pt}
		\begin{tabular}{rl|ccc}
			\tablectop{2-5}
			&  \multirow{2}{*}{\textbf{Dataset}} & \multicolumn{3}{c}{\textbf{Anomaly type}} \\
			&   & \textbf{Object}     & \textbf{Style}      & \textbf{Local}      \\ \cline{2-5} 
			\groupcell{3}{1.7cm}{1.55cm}{\hfill \scriptsize Obj.classif}  & MNIST       & \checkmark &  \crossmark          & \crossmark  \\
			& F-MNIST      & \checkmark &  \crossmark   & \crossmark  \\
			& CIFAR-100     & \checkmark &  \crossmark & \crossmark \\ \cline{2-5} 
			\groupcell{2}{1.7cm}{1.55cm}{\hfill \scriptsize Fine-grained} & Caltech-Birds & \checkmark & \checkmark & \crossmark \\
			& FounderType   & \crossmark & \checkmark &  \crossmark \\ \cline{2-5} 
			\groupcell{1}{1.7cm}{1.55cm}{\hfill \scriptsize Anti-spoofing} & SiW-M         & \checkmark & \checkmark & \checkmark \\
			\tablecbottom{2-5}
		\end{tabular}
	\end{center}
	\caption{\label{tab:db-sum}Summary of evaluation datasets.}
\end{table}

\begin{table*}[!tp]
	\begin{center}
		\hspace{-25pt}
	\begin{tabular}{rl||ccc|ccc}
		\tablectop{2-8}
		&Model                  & CIFAR-100      & MNIST          & F-MNIST        & Caltech-Birds 200 & Fonts           & SiW-M           \\
		\cline{2-8}
		\groupcell{3}{1.9cm}{1.7cm}{\flushright Semi-Supervised} &Deep-SAD (75\%) \cite{deepsad2020} & 88.7           & \underline{99.9}       & \underline{98.1}       & 73.6              & \underline{99.8}        & 85.4            \\
		&Deep-SAD (25\%)  & 87.9           & 98.5           & 95.4           & 70.9              & 99.4            & 76.0            \\
		&Deep-SAD (10\%) & \underline{89.1}       & 96.5           & 88.2           & 66.1              & 98.0            & 80.6            \\
		\cline{2-8}
		\groupcell{9}{1.9cm}{1.7cm}{\hfill One-class} & ADGAN \cite{A:deecke2018image}       & 54.7        & 94.7          & 88.4 & -     & -             & -                  \\
		& GANomaly \cite{akcay2018ganomaly} & 56.5 & 92.8 & 80.9 & - & - & - \\
		& ARNet \cite{Fei2020} &  78.8 & 98.3 & 93.9 & - & - & - \\
		\cdashline{2-8}
		& OCSVM \cite{Schlkopf1999SupportVM}  & -        & 84.7          & 74.2 & 76.3     & -             & -                  \\
		& IF \cite{isolationforest2009} & - & 87.1 & 84.0 & 74.2 & - & - \\
		& PIAD \cite{Tuluptceva2019PerceptualIA}            & 78.8           & \textbf{98.1} & \textbf{94.3} & 63.5              & -               & 81.2           \\
		&GeoTrans \cite{Golan2018DeepAD}         & 84.7           & 96.9           & 92.6           & 66.6              & 92.3            & 81.1            \\
		& MHRot \cite{Hendrycks2019UsingSL}           & 83.6           & 95.2           & 92.5           & 77.6              & 96.7  & 83.1            \\
		\cdashline{2-8}
		&Ours            & \textbf{85.8} &        96.0        &      92.8          & \underline{\textbf{83.2}}   &      \textbf{96.9}           & \underline{\textbf{88.4}} \\
		\tablecbottom{2-8}
	\end{tabular}
	\end{center}
	\caption{\label{tab:sota-table}Comparison with the state-of-the-art AUROC over several datasets, underline indicates best result, bold indicates best one-class learning result. We re-implemented all the methods except the three one-class methods in the first block {\scriptsize (results are from original papers \cite{ A:deecke2018image, Fei2020})}.}
\end{table*}

\begin{table*}[!htbp]
	\begin{center}
	\begin{tabular}{l|cccccccccc|c}
		\tabletop
		Model    & Airplane & Automobile & Bird & Cat  & Deer & Dog  & Frog & Horse & Ship & Truck & Avg  \\ \hline
		VAE \cite{Kingma2014}      & 70.0     & 38.6       & 67.9 & 53.5 & 74.8 & 52.3 & 68.7 & 49.3  & 69.6 & 38.6  & 58.3 \\
		OCSVM \cite{Schlkopf1999SupportVM}    & 63.0     & 44.0       & 64.9 & 48.7 & 73.5 & 50.0 & 72.5 & 53.3  & 64.9 & 50.8  & 58.5 \\
		AnoGAN \cite{schlegl2017anogan}  & 67.1     & 54.7       & 52.9 & 54.5 & 65.1 & 60.3 & 58.5 & 62.5  & 75.8 & 66.5  & 61.8 \\
		PixelCNN \cite{van2016pixcnn} & 53.1     & 99.5       & 47.6 & 51.7 & 73.9 & 54.2 & 59.2 & 78.9  & 34.0 & 66.2  & 61.8 \\
		Deep-SVDD \cite{ruff18deepSvdd}   & 61.7     & 65.9       & 50.8 & 59.1 & 60.9 & 65.7 & 67.7 & 67.3  & 75.9 & 73.1  & 64.8 \\
		OCGAN \cite{perera2019ocgan}   & 75.7     & 53.1       & 64.0 & 62.0 & 72.3 & 62.0 & 72.3 & 57.5  & 82.0 & 55.4  & 65.6 \\
		Puzzle-AE \cite{DBLP:journals/corr/abs-2008-12959} & 78.9 & 78.0 & 69.9 & 54.8 & 75.4 & 66.0 & 74.7 & 73.3 & 83.3 & 69.9 & 72.4 \\
		DROCC \cite{goyal2020DROCC}   & 81.7     & 76.7       & 66.7 & 67.1 & 73.6 & 74.4 & 74.4 & 71.4  & 80.0 & 76.2  & 74.2 \\
		GeoTrans \cite{Golan2018DeepAD}  & 74.7     & 95.7       & 78.1 & 72.4 & 87.8 & 87.8 & 83.4 & 95.5  & 93.3 & 91.3  & 86.0 \\
		\cdashline{1-12}
		Ours     & 75.1         & 96.3           & 84.8     & 74.2     & 91.1     & 89.9     & 88.7     &  95.5     &  94.7    &  91.9     & \textbf{88.2} \\
		\tablebottom
	\end{tabular}
	\end{center}
	\caption{\label{tab:sota-table3}Detailed comparison with one-class state-of-the-art AUROC on CIFAR-10 dataset.}
\end{table*}

Thus we propose to use fine-grained classification datasets in the same one-vs-all protocol. Since discrimination between theses classes is mostly based on local and fine patterns, we can have a good coverage of style anomalies and local anomalies. Also we note that because of the increased shift in object recognition toward fined-grain classification, such datasets have become readily available. For one-vs-all datasets, we used \textbf{MNIST} \cite{726791}, \textbf{Fashion MNIST} \cite{Xiao2017FashionMNISTAN}, \textbf{CIFAR-100} \cite{Krizhevsky2009LearningML}. For the fine-grained dataset, we chose the \textbf{Caltech-Birds 200} database \cite{WelinderEtal2010}.

We also put forward datasets from real anomaly detection problems over different fields. First, we use font recognition challenges as they provide shape-focused style anomaly detection. Indeed two different fonts represent the same characters albeit with a distinctive style. Even though these images lie on a low dimensional manifold compared to natural images, they still provide insight into how well the model can capture small shape hints. In particular, we use \textbf{FounderType-200} \cite{10.1109/CVPR.2017.439} introduced for novelty detection and containing 6700 images per font.
Furthermore, we choose a dataset from face anti-spoofing, where the goal is to discriminate real faces from fake representations of someone's face. Due to the richness and high variability of such frauds, this problem effectively encompasses all three types of anomalies. We use here the \textbf{Spoof in the Wild Multiple} (SiW-M) \cite{dtlfz-2019} database which contains more than 1600 short videos of real faces and presentation attacks. There are 493 real identities along with several types of attacks: paper print, screen replay, masks and partial attacks where only a localized area of the face is fake. The masks are composed of half-masks, paper masks, silicone mask and transparent masks. All evaluation datasets are summarized in Table \ref{tab:db-sum}.

We additionally evaluate our anomaly detection model on out-of-distribution (OOD) protocol. OOD detection, which is broader than anomaly detection, aims at discriminating the training dataset from other data distributions. The "normal" distribution in OOD is therefore usually more diverse and highly multi-modal. We also have a greater overlap in term of class between the in-distribution samples and out-of-distribution samples compared to anomaly detection. Nevertheless it gives us great insight into the multi-modality limits of our model. The most common evaluation setup is to discriminate one training multi-class dataset from other datasets. Here we choose to learn on CIFAR-10 and discriminate CIFAR-100 and the easier Street View House Numbers (\textbf{SVHN}) dataset \cite{Netzer2011}.

For all of the evaluations, the metric used is the area under the ROC curve (\textbf{AUROC}), averaged over all possible normal classes in the case of one-vs-all datasets.

\subsection{Ablation study}

%\iffalse

%\fi

We evaluate in Table \ref{tab:abl-table1} how combining the two tasks of geometric transformation recognition and jigsaw puzzle improves the anomaly detection. We drastically improve performances with a relative error reduction regarding AUROC of 13\% on CIFAR-100, 25\% on Caltech-Birds 200 and 31\% on SIW-M. This validates our statement in Section \ref{sec:method}: the finer the differences between anomaly and normal class, the greater the improvement is by adding the jigsaw task. 

\begin{table}[!htbp]
	\begin{center}
	\begin{tabular}{l|ccc}
		\tabletop
		Auxiliary Task                                              & CIFAR-100     & Caltech-Birds & SiW-M         \\ \hline
		Geometric (G)                                                   & 83.6          & 77.6              & 83.1          \\
		Jigsaw (J)                                                      & 80.1          & 78.5              & 76.3          \\
		Ours (G+J)
		%\begin{tabular}[c]{@{}l@{}}Geometric+\\ Jigsaw\end{tabular}
		& \textbf{85.8} & \textbf{83.2}     & \textbf{88.4} \\
		\tablebottom
	\end{tabular}
	\end{center}
	\caption{\label{tab:abl-table1}AUROC for different tasks, best result is in bold.}
\end{table}

\subsection{Comparison to the state-of-the-art}

We compare our method with different one-class learning state-of-the-art approaches to anomaly detection: reconstruction error generative models with the \textbf{PIAD} model \cite{Tuluptceva2019PerceptualIA}, self-supervised methods with \textbf{GeoTrans} \cite{Golan2018DeepAD} and \textbf{MHRot} \cite{Hendrycks2019UsingSL}. As an addition, we include a semi-supervised learning anomaly detection method \textbf{DeepSAD} \cite{deepsad2020}, which has access to a portion of the anomalies during training. As such, we train it with the same normal samples but three different ratio of the anomaly subclasses: 10\%, 25\% and 75\%.

For the sake of fair comparison in the same conditions, we take the existing implementations or re-implement each method and evaluate each, except for the ADGAN, GANomaly and ARNet which we reference results from their original papers \cite{ A:deecke2018image, akcay2018ganomaly, Fei2020}. 

The results are gathered in Table \ref{tab:sota-table} and \ref{tab:sota-table3}. First of all, we can see our method generally maintains among the best accuracies on simple object anomaly detection, and even improves it on more challenging datasets such as CIFAR-100. Moreover, it greatly improves fine-grained anomaly detection and outperforms state-of-the-art methods which could not be realistically be used for this problem. We also show that our method, without further tuning, improves anti-spoofing detection performances on SiW-M. Finally, we notice our one-class learning model generally reduces the gap with semi-supervised method, and even outperforms these on Caltech-Birds 200 and SiW-M, even though these take advantage of a significant amount of additional anomalous data.

\begin{table}[!htbp]
	\begin{center}
	\begin{tabular}{l|ccc}
		\tabletop
		Metrics & AUROC         & EER           & \begin{tabular}{c}
			APCER \\ (5\%BPCER)
		\end{tabular} \\ \hline
		MHRot \cite{Hendrycks2019UsingSL} & 83.0          & 21.6          & 77.5           \\
		Ours    & \textbf{88.4} & \textbf{18.7} & \textbf{39.1} \\
		\tablebottom
	\end{tabular}
	\end{center}
	\caption{\label{tab:sota-table2}AUROC, EER and APCER at 5\% BPCER of MHRot and our model with jigsaw task on SiW-M dataset, best result is in bold.}
\end{table}

We compare in Table \ref{tab:sota-table2} our method with the second best self-supervised method MHRot on SiW-M. We use metrics more adapted to face presentation attack detection with equal error rate (\textbf{EER}) and the false acceptance rate for the rate of false reject fixed at 5\% (\textbf{APCER@5\%BPCER}). Our comparison does not include other face anti-spoofing methods since we only use real faces training images while these all use a set of presentation attacks during training. Using our method, the APCER@5\%BPCER drops from 77.5 to 39.1 thus also showing promising usage of anomaly detection methods in fraud detection.

\begin{table}[!htbp]
	\begin{center}
	\begin{tabular}{l|ccc}
		\tabletop
		OOD             & SVHN & \multicolumn{1}{c|}{CIFAR-100} & Avg \\ \hline
		VAE \cite{Kingma2014}         & 2.4  & \multicolumn{1}{c|}{52.8}      & 27.6    \\
		Deep-SVDD  \cite{ruff18deepSvdd}     & 14.5 & \multicolumn{1}{c|}{52.1}      & 33.3    \\
		PixelCNN \cite{van2016pixcnn}  & 15.8 & \multicolumn{1}{c|}{52.4}      & 34.1    \\
		RotNet \cite{Gidaris2018UnsupervisedRL}         & 97.9 & \multicolumn{1}{c|}{81.2}      & 89.5   \\
		Ours            & 98.8 & \multicolumn{1}{c|}{83.4}      & \textbf{91.1}    \\
		\cdashline{1-4}
		CSI \cite{tack2020csi}            & 99.8 & \multicolumn{1}{c|}{89.2}      & \underline{94.5}    \\
		\tablebottom
	\end{tabular}
	\end{center}
	\caption{\label{tab:sota-ood}Comparison with state-of-the-art on the Out-Of-Distribution detection protocol with CIFAR-10 as in-distribution, best result is underlined, best pretext task driven method is in bold. }
\end{table}

Lastly, we compare our model with state-of-the-art on OOD detection in Table \ref{tab:sota-ood}. Although not designed specifically for such complex normal class, we obtain better detection rates than other self-supervised anomaly detection methods with pretext tasks.

\section{Conclusion and Future Work}

In this paper, we investigate the power of multi-task self supervision for anomaly detection and show the limits of simple geometric tasks. In more details, we combine two complementary tasks of jigsaw puzzle and geometric transformation recognition. Through an ablation study, we show that this enables it to learn much complex and finer features and therefore better detect anomalies. Finally, we provide a more comprehensive evaluation protocol than previously used datasets in the anomaly detection literature. It presents more challenging datasets and covers object, style and local anomalies. Our method outperforms state-of-the-art, including a semi-supervised method, on most of the fine-grained datasets.

For future work we could explore the combination of more tasks, including generative tasks (in contrast to discriminative tasks used here). Such tasks could range from re-colorization to image in-painting.

%\vfill\newpage
\setlength{\bibitemsep}{.3\baselineskip}

{\small
\bibliographystyle{ieee}
\bibliography{refs}
}

\end{document}